\documentclass[10pt,twocolumn,letterpaper]{article}

\usepackage{cvpr}
\usepackage{times}
\usepackage{epsfig}
\usepackage{graphicx}
\usepackage{amsmath}
\usepackage{amssymb}
\usepackage{multirow}
\usepackage{booktabs}
\usepackage{enumitem,kantlipsum}

\usepackage[pagebackref=true,breaklinks=true,letterpaper=true,colorlinks,bookmarks=false]{hyperref}

\cvprfinalcopy 

\def\cvprPaperID{1934} 

\ifcvprfinal\fi
\begin{document}
\title{DeepFaceFlow: In-the-wild Dense 3D Facial Motion Estimation}

\author{\parbox{16cm}{\centering
    {\large Mohammad Rami Koujan$^{1,4}$, Anastasios Roussos$^{1,3,4}$, Stefanos Zafeiriou$^{2,4}$}\\
    {\normalsize
    $^{1}$College of Engineering, Mathematics and Physical Sciences, University of Exeter, UK\\  $^{2}$Department of Computing, Imperial College London, UK\\$^{3}$ Institute of Computer Science, Foundation for Research and Technology-Hellas (FORTH-ICS), Greece\\ $^{4}$FaceSoft.io, London, UK\\}}
}

\twocolumn[{%
\renewcommand\twocolumn[1][]{#1}%
\maketitle
}]

\begin{abstract}
   Dense 3D facial motion capture from only monocular in-the-wild pairs of RGB images is a highly challenging problem with numerous applications, ranging from facial expression recognition to facial reenactment. In this work, we propose \textbf{DeepFaceFlow}, a robust, fast, and highly-accurate framework for the dense estimation of 3D non-rigid facial flow between pairs of monocular images. Our \textbf{DeepFaceFlow} framework was trained and tested on two very large-scale facial video datasets, one of them of our own collection and annotation, with the aid of occlusion-aware and 3D-based loss function. We conduct comprehensive experiments probing different aspects of our approach and demonstrating its improved performance against state-of-the-art flow and 3D reconstruction methods. Furthermore, we incorporate our framework in a full-head state-of-the-art facial video synthesis method and demonstrate the ability of our method in better representing and capturing the facial dynamics, resulting in a highly-realistic facial video synthesis. Given registered pairs of images, our framework generates 3D flow maps at $\sim 60$ fps.
\end{abstract}

\section{Introduction}
Optical flow estimation is a challenging computer vision task that has been targeted substantially since the seminal work of Horn and Schunck \cite{horn1981determining}. The amount of effort dedicated for tackling such a problem is largely justified by the potential applications in the field, e.g. 3D facial reconstruction \cite{garg2013dense, koujan2018combining, wang2018dense}, autonomous driving \cite{janai2017computer}, action and expression recognition \cite{simonyan2014two, Koujan2020FER}, human motion and head pose estimation \cite{alldieck2017optical, zhu20033d}, and video-to-video translation \cite{wang2018video,Koujan2020head2head}. While optical flow tracks pixels between consecutive images in the 2D image plane, scene flow, its 3D counterpart, aims at estimating the 3D motion field of scene points at different time steps in the 3 dimensional space. Therefore, scene flow combines two challenges: 1) 3D shape reconstruction, and 2) dense motion estimation. Scene flow estimation, which can be traced back to the work of of vedula et al. \cite{vedula1999three}, is a highly ill-posed problem due to the depth ambiguity and the aperture problem, as well as occlusions and variations of illumination and pose, etc. which are very typical of in-the-wild images. To address all these challenges, the majority of methods in the literature use stereo or RGB-D images and enforce priors on either the smoothness of the reconstructed surfaces and estimated motion fields \cite{basha2013multi, pons2007multi, wedel2008efficient, thakur2018sceneednet}  or the rigidity of the motion \cite{vogel20113d}.   

\begin{figure}[t]
    \centering
    \includegraphics[width=.98\columnwidth]{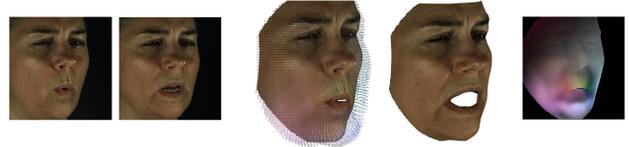}
    \caption{We propose a framework for the high-fidelity 3D flow estimation between a pair of monocular facial images. Left-to-right: \textbf{1 and 2)} input pair of RGB images, \textbf{3)} estimated 3D facial shape of first image rendered with 3D motion vectors from first to second image, \textbf{4)} warped 3D shape of (1) based on estimated 3D flow in (3), \textbf{5)} color-coded 3D flow map of each pixel in (1). For the color coding, see the Supplementary Material.}
    \label{fig:teaser}
\end{figure}
In this work, we seek to estimate the 3D motion field of human faces from in-the-wild pairs of \textbf{monocular} images, see Fig.~\ref{fig:teaser}. 
The output in our method is the same as in scene flow methods, but the fundamental difference is that we use simple RGB images instead of stereo pairs or RGB-D images as input. Furthermore, our method is tailored for human faces instead of arbitrary scenes. For the problem that we are solving, we use the term ``3D face flow estimation''. Our designed framework delivers accurate flow estimation in the 3D world rather than the 2D image space. We focus on the human face and the modelling of its dynamics due to its centrality in myriad of applications, e.g. facial expression recognition, head motion and pose estimation, 3D dense facial reconstruction, full head reenactment, etc. Human facial motion emerges from two main sources: 1) rigid motion due to the head pose variation, and 2) non-rigid motion caused by elicited facial expressions and mouth motions during speech. The reliance on only monocular and \textbf{in-the-wild} images to capture the 3D motion of general objects makes the problem considerably more challenging. Alleviating such obstacles can be made by injecting our prior knowledge about this object, as well as constructing and utilising a large-scale annotated dataset. 
Our contributions in this work can be summarised as follows:
\begin{itemize}[wide, labelwidth=!, labelindent=0pt]
    \item To the best of our knowledge, there does not exist any method that estimates 3D scene flow of \textbf{deformable} scenes using a pair of simple RGB images as input. The proposed approach is the first to solve this problem and this is made possible by focusing on scenes with human faces.
    \item Collection and annotation of a large-scale dataset of human facial videos (more than 12000), which we call \textbf{Face3DVid}. With the help of our proposed model-based formulation, each video was annotated with the per-frame: 1) 68 facial landmarks, 2) dense 3D facial shape mesh, 3) camera parameters, 4) dense 3D flow maps. This dataset will be made publicly available (project's page: \small{\url{https://github.com/mrkoujan/DeepFaceFlow}}).
    \item A robust, fast, deep learning-based and end-to-end framework for the dense high-quality estimation of 3D face flow from only a pair of monocular in-the-wild RGB images. 
    \item We demonstrate both quantitatively and qualitatively the usefulness of our estimated 3D flow in a full-head reenactment experiment, as well as 4D face reconstruction (see supplementary materials).
\end{itemize}
The approach we follow starts from the collection and annotation of a large-scale dataset of facial videos, see section \ref{sec:FaceVid} for details. We employ such a rich dynamic dataset in the training process of our entire framework and initialise the learning procedure with this dataset of in-the-wild videos. Different from other scene flow methods, our framework requires only a pair of monocular RGB images and can be decomposed into two main parts: 1) a shape-initialization network (\textbf{3DMeshReg}) aiming at densely regressing the 3D geometry of the face in the first frame, and 2) a fully convolutional network, termed as \textbf{DeepFaceFlowNet (DFFNet)}, that accepts a pair of RGB frames along with the projected 3D facial shape initialization of the first (reference) frame, provided by \textbf{3DMeshReg}, and produces a dense 3D flow map at the output. 
\section{Related Work}

The most closely-related works in the literature solve the problems of  optical flow and scene flow estimation. Traditionally, one of the most popular approaches to tackle these problems had been through variational frameworks. The work of Horn and Schunck \cite{horn1981determining} pioneered the variational work on \textbf{optical flow}, where they formulated an energy equation with brightness constancy and spatial smoothness terms. Later, a large number of variational approaches with various improvements were put forward \cite{brox2004high, memin1998dense, wedel2009structure, black1996robust, sun2014quantitative}. All of these methods involve dealing with a complex optimisation, rendering them computationally very intensive.
One of the very first attempts for an end-to-end and CNN-based trainable framework capable of estimating the optical flow was made by Dosovitskiy et al. \cite{dosovitskiy2015flownet}. Even though their reported results still fall behind state-of-the-art classical methods, their work shows the bright promises of CNNs in this task and that further investigation is worthwhile. Another attempt with similar results to \cite{dosovitskiy2015flownet} was made by the authors of \cite{ranjan2017optical}. Their framework, called \textbf{SpyNet}, combine a classical spatial-pyramid formulation with deep learning for large motions estimation in a coarse-to-fine approach. As a follow-up method, Ilg et al. \cite{ilg2017flownet} later used the two structures proposed in \cite{dosovitskiy2015flownet} in a stacked pipeline, \textbf{FlowNet2}, for estimating coarse and fine scale details of optical flow, with very competitive performance on the Sintel benchmark. Recently, the authors of \cite{sun2018pwc} put forward a compact and fast CNN model, termed as \textbf{PWC-Net}, that capitalises on pyramidal processing, warping, and cost volumes. They reported the top results on more than one benchmark, namely: MPI Sintel final pass and KITTI 2015. Most of the deep learning-based methods rely on synthetic datasets to train their networks in a supervised fashion, leaving a challenging gap when tested on real in-the-wild images. 

Quite different from optical flow, \textbf{scene flow} methods basically aim at estimating the three dimensional motion vectors of scene points from stereo or RGB-D images. The first attempt to extend optical flow to 3D was made by Vdedula et al. \cite{vedula1999three}. Their work assumed both the structure and the correspondences of the scene are known. Most of the early attempts on scene flow estimation relied on a sequence of stereo images to solve the problem. With the more popularity of depth cameras, more pipelines were utilising RGB-D data as an alternative to stereo images. All these methods follow the classical way of scene flow estimation without using any deep learning techniques or big datasets. The authors of \cite{mayer2016large} led the first effort to use deep learning features to estimate optical flow, disparity, and scene flow from a big dataset. The method of Golyanik et al.  \cite{golyanik2016nrsfm} estimates the 3D flow from a \textbf{sequence} of monocular images, with sufficient diversity in non-rigid deformation and 3D pose, as the method relies heavily on NRSfM. The lack of such diversity, which is common for the type of in-the-wild videos we deal with, could result in degenerate solutions. On the contrary, our method  requires only a \textbf{pair} of monocular images as input. Using only monocular images, Brickwedde et al. \cite{brickwedde2019mono} target dynamic \textbf{street} scenes but impose a \textbf{strong rigidity} assumption on scene objects, making it unsuitable for facial videos.
As opposed to other state-of-the-art approaches, we rely on minimal information to solve the highly ill-posed 3D facial scene flow problem. Given only a pair of monocular RGB images, our novel framework is capable of accurately estimating the 3D flow between them robustly and quickly at a rate of $\sim 60 fps$.

\begin{figure*}[t!]
    \centering
    \includegraphics[width=0.75\linewidth]{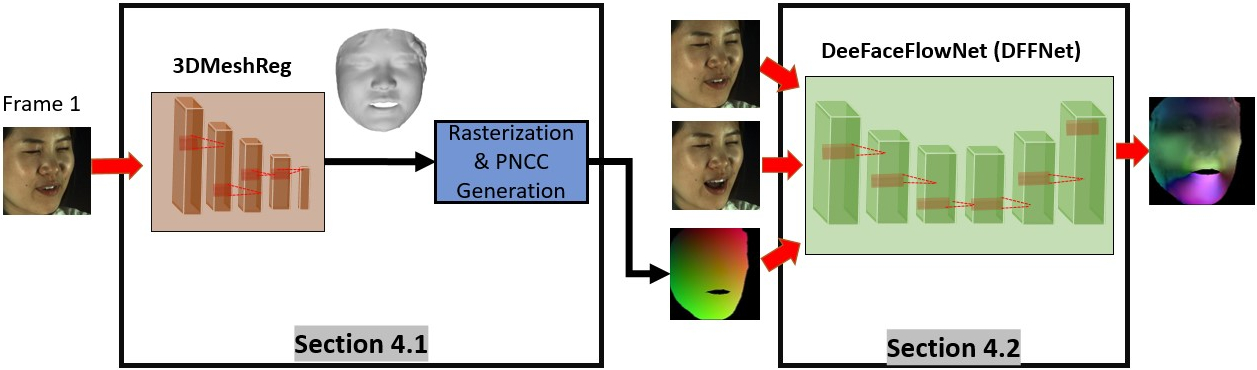}
    \caption{Proposed \textbf{DeepFaceFlow} pipeline for the 3D facial flow estimation. First stage (left): \textbf{3DMeshReg} works as an initialisation for the 3D facial shape in the first frame. This estimation is rasterized in the next step and encoded in an RGB image, termed as Projected Normalised Coordinates Code (PNCC) storing the x-y-z coordinates of each corresponding visible 3D point. Given the pair of images as well as the PNCC, the second stage (right) estimates the 3D flow using a deep fully-convolutional network (\textbf{DeepFaceFlowNet}).}
    \label{fig:pipeline}
\end{figure*}

\section{Dataset Collection and Annotation}
\label{sec:FaceVid}

Given the highly ill-posed nature of the non-rigid 3D facial motion estimation from a pair of monocular images, the size and variability of the training dataset is very crucial \cite{ilg2017flownet, dosovitskiy2015flownet}. 
For this reason, we are based on a large-scale training dataset (\textbf{Face3DVid}), which we construct 
by collecting tens of thousands of facial videos, performing dense 3D face reconstruction on them and then estimating effective pseudo-ground truth of 3D flow maps. 

\subsection{3D Face Reconstruction From Videos}
\label{subsec: 3DMM}

First of all, we \textbf{model the 3D face geometry} using 3DMMs and an additive combination of identity and expression variation. 
This is similar to several recent methods, e.g.~ \cite{Zafeiriou2017,Deng2018,koujan2018combining,gecer2019ganfit}. 
In more detail, 
let $\mathbf{x}=[x_{1}, y_{1}, z_{1}, ..., x_{N}, y_{N}, z_{N}]^T \in \mathbb{R}^{3N}$ be the vectorized form of any 3D facial shape consisting of $N$ 3D vertices. We consider that $\mathbf{x}$ can be represented as:
\begin{equation}
\label{eq:3DMM}
\mathbf{x}(\mathbf{i},\mathbf{e})=\bar{\mathbf{x}}+\mathbf{U}_{id} \mathbf{i}+ \mathbf{U}_{exp} \mathbf{e}
\end{equation}
where $\mathbf{\bar{x}}$ is the overall mean shape vector, $\mathbf{U}_{id} \in \mathbb{R}^{3N\times n_i}$ is the identity basis with $n_i=157$ principal components ($n_i\ll3N$), $\mathbf{U}_{exp} \in \mathbb{R}^{3N\times n_e}$ is the expression basis with  $n_e=28$ principal components ($n_e\ll3N$), and $\mathbf{i} \in \mathbb{R}^{n_i}$, $\mathbf{e} \in \mathbb{R}^{n_e}$ are the identity and expression parameters respectively. The identity part of the model originates from the Large Scale Face Model (LSFM) \cite{booth2018large} and the expression part originates from the work of Zafeiriou et al.~\cite{Zafeiriou2017}. 



To create effective pseudo-ground truth on tens of thousands of videos, we need to perform \textbf{3D face reconstruction} that is both efficient and accurate. 
For this reason, we choose to fit the adopted 3DMM model on the sequence of facial landmarks over each video.  Since this process is done only during training, we are not constrained by the need of online performance. Therefore, similarly to \cite{Deng2018}, we adopt a batch approach that takes into account the information from all video frames simultaneously and exploits the rich dynamic information usually contained in facial videos. It is an energy minimization to fit the combined identity and expression 3DMM model on facial landmarks from all frames of the input video simultaneously. More details are given in the Supplementary Material.

\subsection{Application on a Large-scale  Videos Dataset}
\label{subsec: dataset}

To create our large scale training dataset, we  start from a collection of 12,000 RGB videos with 19 million frames in total and 2,500 unique identities. We apply the 3D face reconstruction method outlined  in Sec.~\ref{subsec: 3DMM}, together with some video pruning steps to omit cases where the automatic estimations had failed. Our final training set consists of videos of our collection that survived the steps of video pruning: \textbf{9750 videos} (81.25\% of the initial dataset) with \textbf{1600 different identities} and around \textbf{12.5M frames}. For more details and exemplar visualisations, please refer to the Supplementary Material.

\subsection{Creation of 3D Flow Annotations}
\label{subsec:flow-annotation}
Given a pair of images $\mathcal{I}_1$ and $\mathcal{I}_2$, coming from a video in our dataset, and their corresponding 3D shapes $\mathbf{S}_1, \mathbf{S}_2$ and pose parameters $\mathbf{R}_1$, $\mathbf{t3d}_1, \mathbf{R}_2, \mathbf{t3d}_2$, the 3D flow map of this pair is created as follows:
\begin{equation}
    \label{eq:flow-computation}
\begin{split}
     \underset{(x,y)\in \mathit{M}}{F(x, y)}&= \underset{t^{j}\in \{\mathit{T}| t^{j} \text{ is visible from pixel (x,y) in} \mathcal{I}_1 \}}{f_{c2}\cdot (\mathbf{R}_2 [\mathbf{S}_2(t^{j}_{1}), \mathbf{S}_2(t^{j}_{2}), \mathbf{S}_2(t^{j}_{3})]} \mathbf{b} \!+\! \mathbf{t}_{3d_2})
     \\-&\underset{t^{j}\in \{\mathit{T}| t^{j} \text{ is visible from pixel (x,y) in} \mathcal{I}_1  \}}{f_{c1}\cdot  (\mathbf{R}_1 [\mathbf{S}_1(t^{j}_{1}), \mathbf{S}_1(t^{j}_{2}), \mathbf{S}_1(t^{j}_{3})]} \mathbf{b} \!+\! \mathbf{t}_{3d_1}),
\end{split}
\end{equation}
where $\mathit{M}$ is the set of foreground pixels in $\mathcal{I}_{1}$, $\mathbf{S}\in \mathbb{R}^{3\times N}$ is the matrix storing the column-wise x-y-z coordinates of the  $N$-vertices 3D shape of $\mathcal{I}_{1}$, $\mathbf{R}\in \mathbb{R}^{3\times3}$ is the rotation matrix, $\mathbf{t}_{3d}\in \mathbb{R}^{3}$ is the 3D translation, $f_{c1}$ and $f_{c2}$ are the scales of the orthographic camera for the first and second image, respectively, $t^{j}=[t^{j}_{1}, t^{j}_{2}, t^{j}_{3}]$ ($t^{j}_{i}\in \{1, .., N\}$) is the visible triangle from pixel $(x, y)$ in image $\mathcal{I}_1$ detected by our hardware-based renderer, $\mathit{T}$ is the set of all triangles composing the mesh of $\mathbf{S}$, and $\mathbf{b}\in \mathbb{R}^{3}$ is the barycentric coordinates of pixel $(x,y)$ lying inside the projected triangle $t^{j}$ on image $\mathcal{I}_1$. All background pixels in equation \ref{eq:flow-computation} are set to zero and ignored during training with the help of a masked loss. It is evident from equation \ref{eq:flow-computation} that we do not care about the visible vertices in the second frame and only track in 3D those were visible in image $\mathcal{I}_1$ to produce the 3D flow vectors. Additionally, with this flow representation, the x-y coordinates alone of the 3D flow map ($F(x, y)$) designate the 2D optical flow components in the image space directly.
\vspace{-.3cm}
\section{Proposed Framework}
\label{sec:framework}
Our overall designed framework is demonstrated in figure \ref{fig:pipeline}. We expect as input two RGB images $\mathcal{I}_{1}, \mathcal{I}_{2}\in \mathbb{R}^{W\times H\times 3}$ and produce at the output an image $F\in \mathbb{R}^{W\times H\times 3}$ encoding the per-pixel 3D optical flow from $\mathcal{I}_{1}$ to $\mathcal{I}_{2}$. The designed framework is marked by two main stages: 1) \textbf{3DMeshReg}: 3D shape initialisation and encoding of the reference frame $\mathcal{I}_{1}$, 2) \textbf{DeepFaceFlowNet (DFFNet)}: 3D face flow prediction. The entire framework was trained in a supervised manner, utilising the collected and annotated dataset, see section \ref{subsec: dataset}, and fine-tuned on the 4DFAB dataset \cite{cheng20184dfab}, after registering the sequence of scans coming from each video in this dataset to our 3D template. Input frames were registered to a 2D template of size $224\times224$ with the help of the 68 mark-up extracted using \cite{guo2018stacked} and fed to our framework.

\subsection{3D Shape Initialisation and Encoding}
To robustly estimate the per-pixel 3D flow between a pair of images, we provide the \textbf{DFFNet} network, section \ref{subsec: flow-estimation}, not only with $\mathcal{I}_1\&\mathcal{I}_2$, but also with another image that stores a  \textit{Projected Normalised coordinates Code} ($\mathcal{PNCC}$) of the estimated 3D shape of the reference frame $\mathcal{I}1$, i.e. $\mathcal{PNCC}\in \mathbb{R}^{W\times H\times 3}$. The $\mathcal{PNCC}$ codes that we consider are essentially images encoding the normalised x,y, and z coordinates of facial vertices visible from each corresponding pixel in $\mathcal{I}_{1}$ based on the camera's view angle. The inclusion of such images  allows the CNN to better associate each RGB value in $\mathcal{I}_{1}$ with the corresponding point in the 3D space, providing the network with a better initialisation in the problem space and establishing a reference 3D mesh that facilitates the warping in the 3D space during the course of training. Equation \ref{eq: PNCC} shows how to compute the $\mathcal{PNCC}$ image of the reference frame $\mathcal{I}_{1}$.
\begin{equation}
\label{eq: PNCC}
\begin{split}
     \underset{(x,y)\in \mathit{M}}{\mathcal{PNCC}(x, y)}= \mathcal{V}(\mathbf{S}, \mathbf{c})= \underset{t^{j}\in \{\mathit{T}| t^{j} \text{ is visible from} (x,y)\}}{\mathbf{P} (\mathbf{R} [\mathbf{S}(t^{j}_{1}), \mathbf{S}(t^{j}_{2}), \mathbf{S}(t^{j}_{3})]} \mathbf{b}+ \mathbf{t}_{3d}),
\end{split}
\end{equation}
where $\mathcal{V}(., .)$ is the function rendering the normalised version of $\mathbf{S}$, $\mathbf{c}\in \mathbb{R}^{7}$ is the camera parameters, i.e. rotation angles, translation and scale ($\mathbf{R}, \mathbf{t}_{3d}, f_{c}$), and $\mathbf{P}$ is a $3\times 3$ diagonal matrix with main diagonal elements ($\frac{f_c}{W}, \frac{f_c}{H}, \frac{f_c}{D}$). The multiplication with $\mathbf{P}$ scales the posed 3D face with $f_c$ to be first in the image space coordinates and then normalises it with the width and height of the rendered image size and the maximum z value D computed from the entire dataset of our annotated 3D shapes. This results in an image with RGB channels storing the normalised ([0, 1]) x-y-z coordinates of the corresponding rendered 3D shape. The rest of the parameters in equation \ref{eq: PNCC} are detailed in section \ref{subsec:flow-annotation} and utilised in equation \ref{eq:flow-computation}.


\textbf{3DMeshReg.} The $\mathcal{PNCC}$ image generation discussed in equation \ref{eq: PNCC} still lacks the estimation of the 3D facial shape $\mathbf{S}$ of $\mathcal{I}_{1}$. We deal with this problem by training a deep CNN, termed as \textbf{3DMeshReg}, that aims at regressing a dense 3D mesh $\mathbf{S}$ through per-vertex 3D coordinates estimations. We use our collected dataset (\textbf{Face3DVid}) and the 4DFAB \cite{cheng20184dfab} 3D scans to train this network in a supervised manner. We formulate a loss function composed of two terms:
\begin{equation}
    \label{eq: 3DMehsReg}
    \begin{split}
        \mathcal{L}(\Phi)= \dfrac{1}{N}\sum\limits_{i=1}^{N}|| \mathbf{s}_{i}^{GT}-\mathbf{s}_{i}||^2+ \dfrac{1}{O}\sum\limits_{j=1}^{O}|| \mathbf{e}_{j}^{GT}-\mathbf{e}_{j}||^2.
    \end{split}
\end{equation}
The first term in the above equation penalises the deviation of each vertex 3D coordinates from the corresponding ground-truth vertex ($\mathbf{s}_i=[x, y, z]^T$), while the second term ensures similar edge lengths between vertices in the estimated and ground-truth mesh, given that $\mathbf{e}_{j}$ is the $\ell_2$ distance between vertices $\mathbf{v}^{1}_{j}$ and $\mathbf{v}^{2}_{j}$ defining edge $j$ in the original ground-truth 3D template. Instead of estimating the camera parameters $\mathbf{c}$ separately, which are needed at the very input of the renderer, we assume a Scaled Orthographic Projection (SOP) as the camera model and train the network to regress directly the scaled 3D mesh by multiplying the x-y-z coordinates of each vertex of frame $i$ with $f_{c}^{i}$. 


\begin{figure*}[t!]
    \centering
    \includegraphics[width=0.75\linewidth, height=0.4\linewidth]{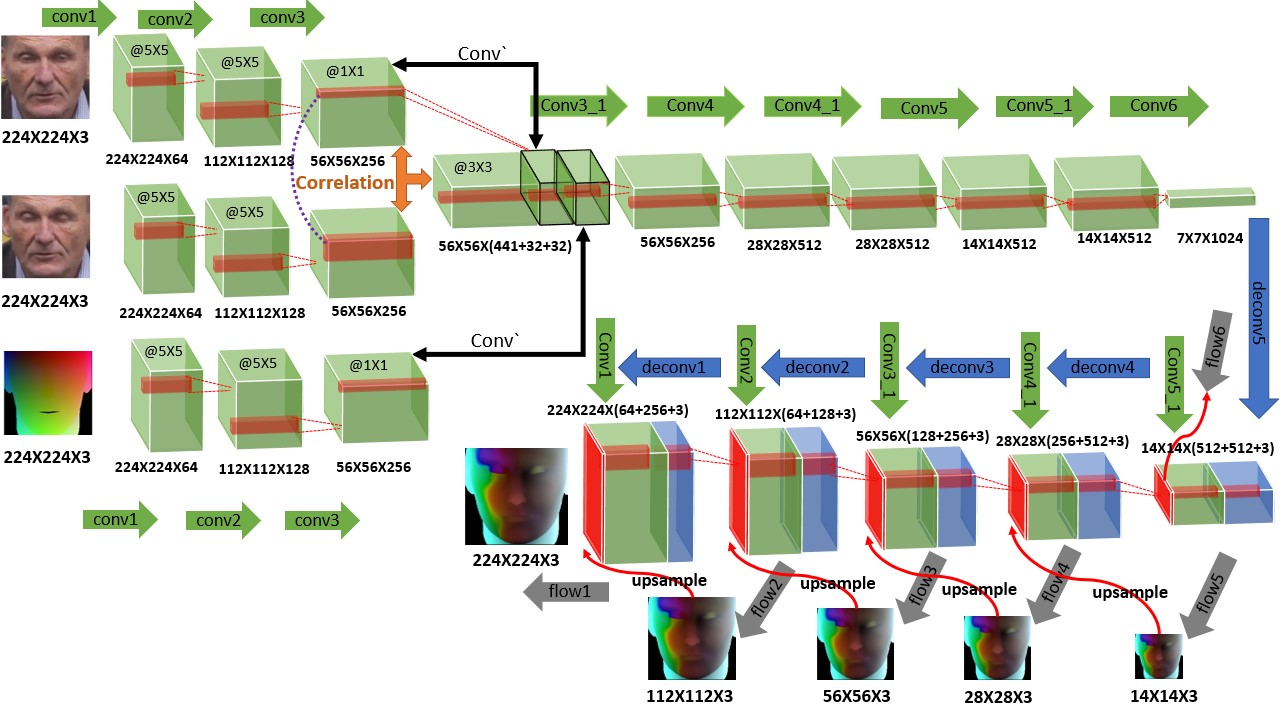}
    \caption{Architecture of our designed \textbf{DFFNet} for the purpose of estimating the 3D flow between a pair of RGB images. }
    \label{fig:FaceFlowNet}
\end{figure*}

\subsection{Face Flow Prediction}
\label{subsec: flow-estimation}
Given $\mathcal{I}_{1}$, $\mathcal{I}_{2}$ and $\mathcal{PNCC}$ images, the 3D flow estimation problem is a mapping $\mathcal{F}: \{\mathcal{I}_{1}, \mathcal{I}_{2}, \mathcal{PNCC}\}\in \mathbb{R}^{W\times H\times 9}\rightarrow F^{W\times H\times 3}$. Using both the annotated \textbf{Face3DVid} detailed in section \ref{sec:FaceVid} and 4DFAB \cite{cheng20184dfab} datasets, we train a fully convolutional encoder-decoder CNN structure ($\mathcal{F}$), called \textbf{DeepFaceFlowNet (DFFNet)}, that takes three images, namely: $\mathcal{I}_{1}$, $\mathcal{I}_{2}$ and $\mathcal{PNCC}$, and produces the 3D flow estimate from each foreground pixel in $\mathcal{I}_{1}$ to $\mathcal{I}_{2}$ as a $W\times H\times 3$ image. The designed network follows the generic U-Net architecture with skip connections \cite{ronneberger2015u} and was inspired particularly by FlowNetC \cite{dosovitskiy2015flownet}, see figure \ref{fig:FaceFlowNet}. Distinguished from FlowNetC, we extend the network to account for the $\mathcal{PNCC}$ image at the input and modify the structure to account for the 3D flow estimation task, rather than 2D optical flow.  We propose the following two-term  loss function:
\begin{equation}
\scriptstyle
    \mathcal{L}(\Psi)= \sum\limits_{i=1}^{L}w_{i}||F^{GT}_{i}-F_{i}(\Psi)||_\mathbf{F}+ \alpha || \mathcal{I}_{1}- \mathcal{W}(F,\mathcal{PNCC};\mathcal{I}_{2})||_\mathbf{F}^2
    \label{eq:loss-func}
\end{equation}
The first term in Eq.~\eqref{eq:loss-func} is the \textbf{endpoint error}, which corresponds to a 3D extension of the standard error measure for optical flow methods. It computes the frobenius norm ($||. ||_\mathbf{F}$) error between the estimated 3D flow $F(\Psi)$ and the ground truth $F^{GT}$, with $\Psi$ representing the network learnable weights. Practically, since at the decoder part of our \textbf{DFFNet} each  fractionally-strided convolution operation, aka. deconvolution, produces an estimation of the flow at different resolutions, we compare this multi-resolution 3D flow with the downsampled versions of the $F_{GT}$, up until the full resolution at stage $L$, and use the weighted sum of the frobenius norm error as a penalisation term.
The second term in Eq.~\eqref{eq:loss-func} is the 
\textbf{photo-consistency error}, which assumes that the colour of each point does not change from $\mathcal{I}_{1}$ to $\mathcal{I}_{2}$. The warping operation was done with the help of the warping function $\mathcal{W(., .)}$. This function warps the 3D shape of $\mathcal{I}_{1}$ encoded inside the $\mathcal{PNCC}$ image using the estimated flow $F$ and samples $\mathcal{I}_{2}$ at the vertices of the resultant projected 3D shape. The warping function in equation \ref{eq:loss-func} was implemented by a differentiable layer detecting the occlusions by the virtue of our 3D flow, and sampling the second image (backward warping) in a differentiable manner at the output stage of our \textbf{DFFNet}. The scale $\alpha$ is used for the sake of terms balancing while training. 
\vspace{-0.3cm}
\section{Experiments}

In this section, we compare our framework with state-of-the-art methods in optical flow and 3D face reconstruction. We ran all the experiments on an NVIDIA DGX1 machine.

\subsection{Datasets}
Although the collected \textbf{Face3DVid} dataset has a wide variety of facial dynamics and identities captured under plenitude of set-ups and viewpoints depicting the in-the-wild scenarios of videos capture, the dataset was annotated with pseudo ground-truth 3D shapes, not real 3D scans. Relying only on this dataset, therefore, for training our framework could result in mimicking the performance of the 3DMM-based estimation, which we want ideally to initialise with and then depart from. Thus, we fine-tune our framework on the 4DFAB dataset \cite{cheng20184dfab}. The 4DFAB dataset is a large-scale database of dynamic high-resolution 3D faces with subjects displaying both spontaneous and posed facial expressions with the corresponding per-frame 3D scans. We leave a temporal gap between consecutive frames sampled from each video if the average 3D flow per pixel is $<=1$ between each pair. In total, around \textbf{3M} image pairs (1600 subjects) form the \textbf{Face3DVid} dataset and \textbf{500K} from the 4DFAB (175 subjects) were used for training/testing purposes. We split the \textbf{Face3DVid} into training/validation vs test (80$\%$ vs 20$\%$) in the first phase of the training. Likewise, 4DFAB dataset was split into training/validation vs test (80$\%$ vs 20$\%$) during the fine-tuning. 

\subsection{Architectures and Training Details}
\begin{figure}
    \centering
    \includegraphics[scale=0.11]{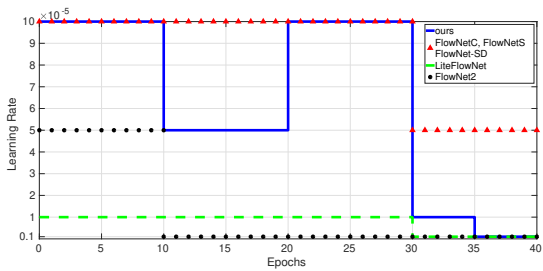}
    \caption{Training schedule for the learning rate used while training our network and other state-of-the-art approaches for 3D flow estimation. The first 20 epochs for all methods were ran on \textbf{Face3DVid} dataset and the next 20 on \textbf{4DFAB}. Each training epoch on \textbf{Face3DVid} and \textbf{4DFAB}  is composed of $18.75\cdot10^4$ and $3.13\cdot10^4$ iterations, respectively, each with batch size of 16.}
    \label{fig:LR_schedule}
\end{figure}

Our pipeline consists of two networks (see Fig.~\ref{fig:pipeline}):
\\
\textbf{a) 3DMeshReg}: The aim of this network is to accept an input image ($\mathcal{I}_{1} \in \mathbb{R}^{224\times 224 \times 3}$ ) and regress the per-vertex (x, y, z) coordinates describing the subject's facial geometry. ResNet50 \cite{he2016deep} network architecture was selected and trained for this purpose after replacing the output fully-connected (\textbf{fc}) layer with a convolutional one ($3\times3, 512$) and then a linear \textbf{fc} layer with $\sim1.5k\times3$ neurons. This network was trained initially and separately from the rest of the framework on the \textbf{Face3DVid} dataset and then fine-tuned on the 4DFAB dataset \cite{cheng20184dfab}. Adam optimizer was used \cite{kingma2014adam} during the training with learning rate of 0.0001, $\beta_{1}=0.9$, $\beta_{2}=0.999$, and batch size 32.
\\
\textbf{b) DFFNet}: Figure \ref{fig:FaceFlowNet} shows the structure of this network. Inspired by FlowNetC \cite{dosovitskiy2015flownet}, this network has similarly nine convolutional layers. The first three layers use kernels of size $5\times5$ and the rest have kernels of size $3\times3$. Where occurs, downsampling is carried out with strides of 2 and non-linearity is implemented with ReLU layers. We extend this architecture at the input stage by a branch dedicated for processing the $\mathcal{PNCC}$ image. The feature map generated at the end of the $\mathcal{PNCC}$ branch is concatenated with the correlation result between the feature maps of $\mathcal{I}_{1}$ and $\mathcal{I}_2$. For the correlation layer, we follow the implementation suggested by \cite{dosovitskiy2015flownet} and we keep the same parameters for this layer (neighborhood search size is 2*(21)+1 pixels). At the decoder section, the flow is estimated from the finest level up until the full resolution. While training, we use a batch size of 16 and the Adam optimization algorithm \cite{kingma2014adam} with the default parameters recommended in \cite{kingma2014adam} ($\beta_{1}=0.9$ and $\beta_{2}=0.999$). Figure \ref{fig:LR_schedule} demonstrates our scheduled learning rates over epochs for training and fine-tuning. We also set $w_i=1$ and $\alpha=10$ in equation \ref{eq:loss-func} and normalise input images to the range [0, 1]. While testing, our entire framework takes only around 17ms (6ms (\textbf{3DMehsReg})+ 6ms (rasterization $\&$PNCC generation) + 5ms (\textbf{DFFNet})) to generate the 3D dense flow map, given a registered pair of images.

 
\subsection{Evaluation of 3D Flow Estimation}

\begin{figure*}[t!]
    \centering
    \includegraphics[trim=0 150 0 0, clip,  width=0.75\linewidth]{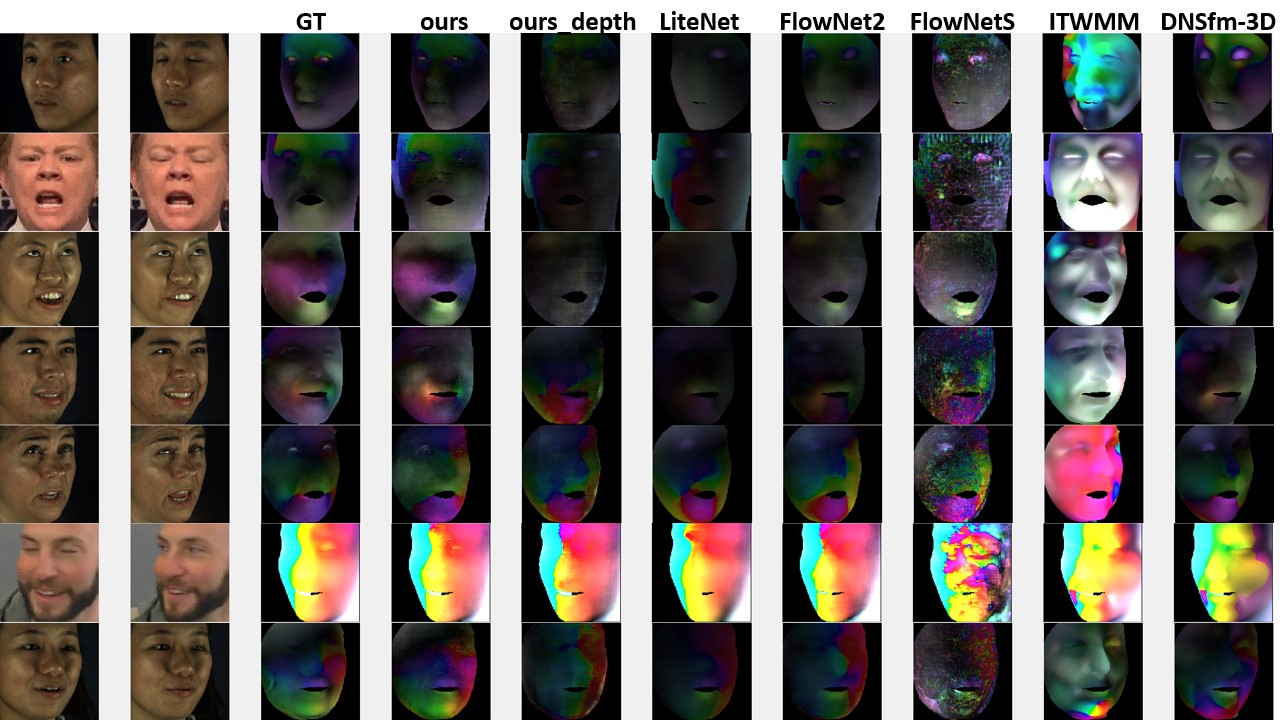}
    \caption{Color-coded 3D flow estimations of random test pairs from the \textbf{Face3DVid} and \textbf{4DFAB} datasets. Left-to-right: pair of input RGB images, Ground Truth, ours, ours$\_$depth, compared methods.  For the color coding, see the Supp.~Material.}
    \label{fig:3Dflow_est_2mf2}
\end{figure*}

In this section, we quantitatively evaluate the ability of our approach in estimating the 3D flow. 
As there exist no other methods for 3D scene flow from simple RGB images, we adapt existing methods that solve closely-related problems so that they produce 3D flow estimation. 
In more detail, we use two 3D reconstruction methods (ITWMM \cite{booth20183d} and DNSfM-3DMM \cite{koujan2018combining}), as well as four optical flow methods after retraining them all specifically for the task of 3D flow estimation. 
The four optical flow methods include the best performing methods in table \ref{table:1} on our datasets (LiteNet$\&$FlowNet2) as well as two additional baselines (FlowNetS$\&$FlowNetC). \\
To estimate the 3D flow by ITWMM and DNSfM-3DMM, we first generate the per-frame dense 3D mesh of each test video by passing a single frame at a time to the ITWMM method and the entire video for the DNSfM-3DMM (as it is a video-based approach). Then, following our annotation procedure discussed in \ref{subsec:flow-annotation}, the 3D flow values for each pair of test images were obtained. \\
Since the deep learning-based methods we compare against in this section were proposed as 2D flow estimators, we modify the sizes of some filters in their original architectures so that their output flow is a 3-channel image storing the x-y-z coordinates of the flow and train them on our 3D-facial-flow datasets with the learning rate schedules reported in figure \ref{fig:LR_schedule}. FlowNet2 is a very deep architecture (around 160M parameters) composed of stacked networks. As suggested in \cite{ilg2017flownet}, we did not train this network in one go, but instead sequentially. More specifically, we fused the separately trained individual networks (FlowNetS, FlowNetC, and FlowNetSD \cite{ilg2017flownet}) on our datasets together and fine-tuned the entire stacked architecture, see \ref{fig:LR_schedule} for the learning rate schedule. Please consult the supplementary material for more information on what we modified exactly in each flow network we compare against here.\\
 Table \ref{table:2} shows the generated facial AEPE results by each method on the \textbf{Face3DVid} and \textbf{4DFAB} datasets. Our proposed architecture and its variant (`ours$\_$depth') report the lowest (best) AEPE numbers on both datasets. Figure \ref{fig:3Dflow_est_2mf2} visualises some color-coded 3D flow results produced by the methods presented in table \ref{table:2}. To color-code the 3D flow, we convert the x-y-z estimated flow coordinates from Cartesian to spherical coordinates and normalise them so that they represent the coordinates of an HSV coloring system, more details on that are available in the supplementary material. It is noteworthy that the 3D facial reconstruction methods we compare against fail to produce as accurate tracking of the 3D flow as our approach. Their result is not smooth and consistent in the model space, resulting in a higher intensity motion in the space. This can be attributed to the fact that such methods pay attention to the fidelity of the reconstruction from the camera's view angle more than the 3D temporal flow. On the other hand, the other deep architectures we train in this section are unable to capture the full facial motion with same precision, with more fading flow around cheeks and forehand.  
 
\begin{table}
\small
\centering
\caption{Comparison between our obtained 3D face flow results against state-of-the-art methods on the test splits of the 4DFAB and \textbf{Face3DVid} datasets. Comparison metric is the standard Average End Point Error (AEPE) }
\begin{tabular}{|c| c c|} 
 \hline
 Method & \textbf{4DFAB}($\downarrow$) &  \textbf{Face3DVid}($\downarrow$) \\ [0.5ex] 
 \hline
 ITWMM \cite{booth20183d}& 3.43  &   4.1   \\
 DNSfM-3DMM \cite{koujan2018combining}& 2.8  & 3.9    \\
 FlowNetS \cite{dosovitskiy2015flownet}& 2.25 &   3.7   \\ 
 FlowNetC \cite{dosovitskiy2015flownet}& 1.95  &2.425     \\
 FlowNet2 \cite{ilg2017flownet}&  1.89& 2.4  \\
 LiteNet \cite{hui2018liteflownet}& 1.5 & 2.2\\
 ours$\_$depth &1.6 &  1.971  \\
 ours &  \textbf{1.3} &  \textbf{1.77}  \\
 \hline
\end{tabular}
\label{table:2}
\end{table}

 \subsection{Evaluation of 2D Flow Estimation}
 
 \begin{figure*}
    \centering
    \includegraphics[trim=0 150 0 0, clip, width=0.75\linewidth]{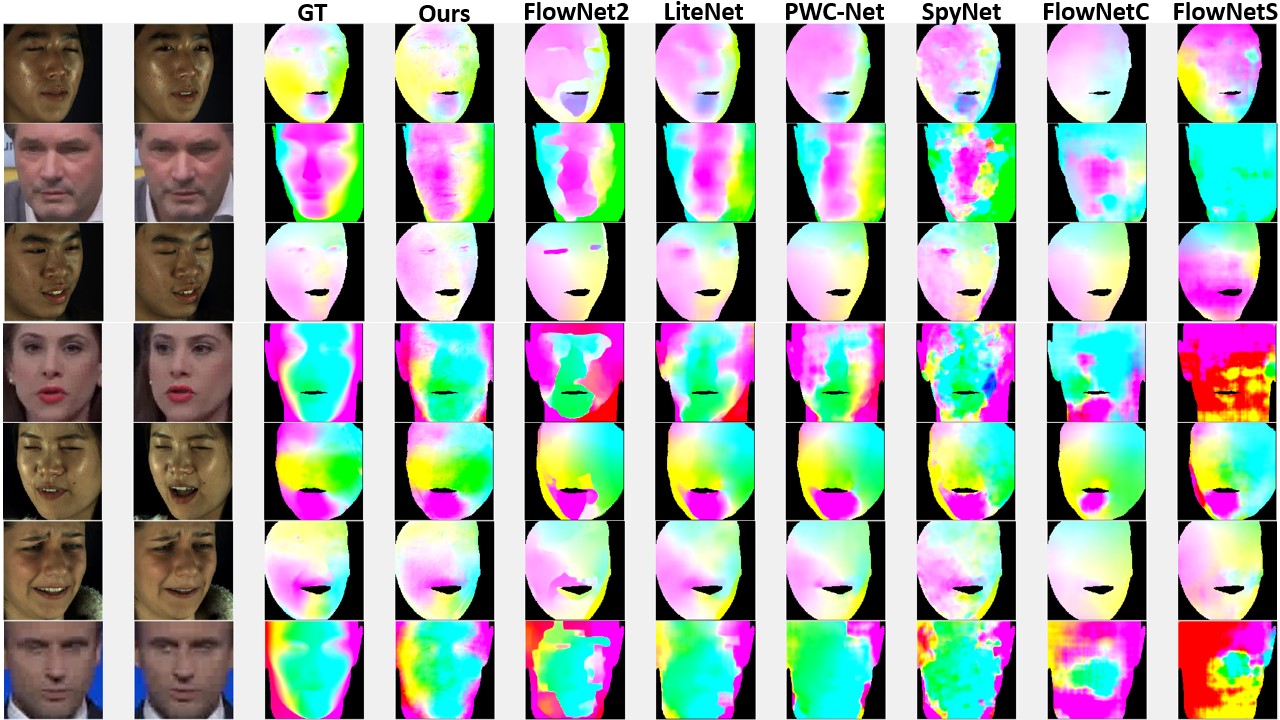}
    \caption{Color-coded 2D flow estimations. Rows are random samples from the test split of the \textbf{Face3DVid} and 4DFAB datasets and their 2D flow estimations. The first two columns of each row show the input pair of RGB images. For the color coding, see the Supp.~Material.}
    \label{fig:2Dflow_estimation}
\end{figure*}

 The aim of this experiment is to probe the performance of our framework in estimating the 2D optical facial flow between a pair of facial images by keeping only the displacements produced at the output in the x and y directions while ignoring those in the z direction. We separate the comparisons in this section into two parts. Firstly, we evaluate our method against generic 2D flow approaches using their best performing trained models provided by the original authors of each. Secondly, we train the same architectures from scratch on the same datasets we train our framework on, namely the training splits of \textbf{Face3DVid} and \textbf{4DFAB}, using a learning rate of 1e-4 that drops 5 times each 10 epochs. We keep the same network design as provided by each paper's authors and only train the network to minimise a masked loss composed of photo-consistency and data terms. The masked loss is computed with the help of a foreground (facial) mask for each reference image provided with our utilised datasets. Table \ref{table:1} presents the obtained facial Average End Point Error (AEPE) metric values by our proposed approach against other state-of-the-art optical flow prediction methods on the test splits of \textbf{Face3DVid} and \textbf{4DFAB} datasets. As can be noted from table \ref{table:1}, our proposed method always achieves the smallest (best) AEPE values on both employed datasets. As expected, the AEPE values decrease when training the other methods on our dataset for the specific task of facial 2D flow estimation. However, our method still produces lower errors and outperforms the compared against methods on this task. The `ours$\_$depth' variant of our network comes as the second best performing method on both datasets. This variant was trained in a very similar manner to our original framework but with feeding the \textbf{DFFNet} with $\mathcal{I}_{1}$, $\mathcal{I}_{2}$ and only the $z$ coordinates (last channel) of the $\mathcal{PNCC}$ image while ignoring the $x$ and $y$ (first two channles). Figure \ref{fig:2Dflow_estimation} demonstrates some qualitative results generated by the methods reported in table \ref{table:1}, as well as ours. Please refer to the supplementary material for more information on the color-coding followed for encoding the flow values.

\begin{table}
\small
\centering
\caption{Comparison between our obtained 2D flow results against state-of-the-art methods on the test splits of the 4DFAB and \textbf{Face3DVid} datasets. Comparison metric is the standard Average End Point Error (AEPE). `original models' refers to trained models provided by the authros of each, and `trained from scratch' indicates that the same architectures were trained on the training sets of both \textbf{Face3DVid} and \textbf{4DFAB} to estimate the 2D facial flow.}
\begin{tabular}{|p{1.9cm}| p{0.75cm} c| p{0.8cm} c|} 
\hline
  \multirow{2}{*}{Method}& \multicolumn{2}{|c|}{original models} & \multicolumn{2}{|p{2.8cm}|}{trained from scratch} \\
  \cline{2-5}
  & \textbf{4DFAB} &  \textbf{Face3DVid} & \textbf{4DFAB} & \textbf{Face3DVid}\\ 
 \hline
 FlowNetS \cite{dosovitskiy2015flownet}& 1.832 &   5.1425 &  1.956& 2.6  \\ 
 SpyNet \cite{ranjan2017optical}&1.31  &3 &  1.042& 1.5\\
 FlowNetC \cite{dosovitskiy2015flownet}&  1.212 &  2.6 &  1.061&  1.498\\
 UnFlow \cite{meister2018unflow}&1.163  & 2.6553 &  1.055& 1.45\\
 LiteNet \cite{hui2018liteflownet}& 1.16 & 2.6&  1.018&1.268\\
 PWC-Net\cite{sun2018pwc}&1.159  &  2.625 & 1.035 &1.371\\
 FlowNet2 \cite{ilg2017flownet} &  1.15 &   2.6187 & 1.063 & 1.352 \\
 ours$\_$depth & 0.99 & 1.176 & 0.99 &   1.176   \\
 ours &  \textbf{0.941} &  \textbf{1.096}&  \textbf{0.941}&  \textbf{1.096}\\
 \hline
\end{tabular}
\label{table:1}
\end{table}

\subsection{Video-to-Video Synthesis With 3D Flow}
\begin{figure}
    \centering
    \includegraphics[scale=0.4]{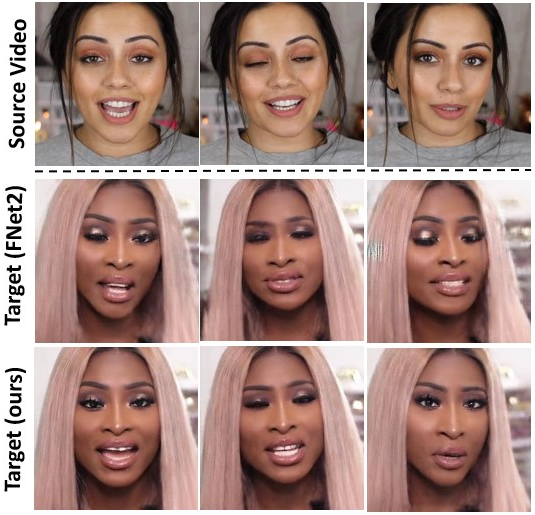}
    \caption{Full-head reenactment using \cite{wang2018video} combined with either FlowNet2 (second row) or our 3D flow approach (last row).}
    \label{fig:reenactment}
\end{figure}

We further investigate the functionality of our proposed framework in capturing the human facial 3D motion and successfully employing it in a full-head reenactment application. Towards that aim, we use the recently proposed method of \cite{wang2018video}, which is in essence a general video-to-video synthesis approach mapping a source (conditioning) video to a photo-realistic output one. The authors of \cite{wang2018video} train their framework in an adversarial manner and learn the temporal dynamics of a target video during the training time with the help of the 2D flow estimated by \textbf{FlowNet2} \cite{ilg2017flownet}. In this experiment, we replace the \textbf{FlowNet2} employed in \cite{wang2018video} by our proposed approach and aid the generator and video discriminator to learn the temporal facial dynamics represented by our 3D facial flow.
We firstly conduct a self-reenactment test as done in \cite{wang2018video}, where we divide each video into a train/test splits (first 2 third vs last third) and report the average per-pixel RGB error between fake and real test frames. Table \ref{table:3} reveals the average pixel distance obtained for 4 different videos we trained a separate model for each. The only difference between the second and third row of table \ref{table:3} is the flow estimation method, everything else (structure, loss functions, conditioning, etc.) is the same. As can be noted from table \ref{table:3}, our 3D flow better reveals the facial temporal dynamics of the training subject and assists the video synthesis generator in capturing these temporal characteristics, resulting in a lower error. In the second experiment, we make a full-head reenactment test to fully transfer the head pose and expression from the source person to a target one. Figure \ref{fig:reenactment} manifests the synthesised frames using our 3D flow and the 2D flow of \textbf{FlowNet2}. Looking closely at figure \ref{fig:reenactment}, our 3D flow results in a more photo-realistic video synthesis with highly accurate head pose, facial expression, as well as temporal dynamics, while the manipulated frames generated with \textbf{FlowNet2} fail to demonstrate the same fidelity. More details regarding this experiment are in the supplementary material.


\begin{table}[h]
  \small
  \centering
  \caption{Average RGB distance obtained under a self-reenactment setup on 4 videos (each with 1K test frames) using either FlowNet2 \cite{ilg2017flownet} or our facial 3D flow with the method of Wang et al. \cite{wang2018video}}
    \begin{tabular}{c|c|c|c|c}
     \textbf{Video} &  1 &  2 & 3 & 4 \\
     \hline
     \cite{wang2018video}$+$\textbf{FlowNet2} ($\downarrow$)& 7.5& 9.5& 8.7 &9.2\\
     \hline
     \cite{wang2018video}$+$\textbf{Ours (3D flow)} ($\downarrow$) &\textbf{6.3}&    \textbf{7.9}&   \textbf{7.5}&    \textbf{7.7}\\
    \end{tabular}
  \label{table:3}
\end{table}

\vspace{-.5cm}
\section{Conclusion and Future Work}
In this work, we put forward a novel and fast framework for densely estimating the 3D flow of human faces from only a pair of monocular RGB images. The framework was trained on a very large-scale dataset of in-the-wild facial videos (\textbf{Face3DVid}) and fine-tuned on a 4D facial expression database (4DFAB \cite{cheng20184dfab}) with ground-truth 3D scans. We conduct extensive experimental evaluations that show that the proposed approach: \textbf{a)} yields highly-accurate estimations of 2D and 3D facial flow from monocular pair of images and successfully captures complex non-rigid motions of the face and \textbf{b)} outperforms many state-of-the-art approaches  in estimating both the 2D and 3D facial flow, even when training other approaches under the same setup and data. We additionally reveal the promising potential of our work in a full-head facial manipulation application that capitalises on our facial flow to produce highly loyal and photo-realistic fake facial dynamics indistinguishable from real ones. 

\section*{Acknowledgement}
Stefanos Zafeiriou acknowledges support from EPSRC Fellowship DEFORM (EP/S010203/1)

{\small
\bibliographystyle{ieee_fullname}
\bibliography{egbib}
}

\end{document}